# Path Planning Algorithm for Extinguishing Forest Fires

M.P.Sivaram Kumar, S.Rajasekaran

**Abstract**— One of the major impacts of climatic changes is due to destroying of forest. Destroying of forest takes place in many ways but the majority of the forest is destroyed due to wild forest fires. In this paper we have presented a path planning algorithm for extinguishing fires which uses Wireless Sensor and Actor Networks (WSANs) for detecting fires. Since most of the works on forest fires are based on Wireless Sensor Networks (WSNs) and a collection of work has been done on coverage, message transmission, deployment of nodes, battery power depletion of sensor nodes in WSNs we focused our work in path planning approach of the Actor to move to the target area where the fire has occurred and extinguish it. An incremental approach is presented in order to determine the successive moves of the Actor to extinguish fire in an environment with and without obstacles. This is done by comparing the moves determined with target location readings obtained using sensors until the Actor reaches the target area to extinguish fires.

**Index Terms**— Forest Fires, Wireless Sensor and Actor Networks, Path Planning

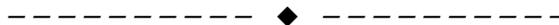

## 1 INTRODUCTION

Nowadays the likelihood of occurrence of forest fire is steadily growing because of change in climate due to cut down of trees in forest for various reasons such as Land for shelter, Land for cultivation and killing of dangerous wildlife animals which pose threat to human lives. The Forest Survey of India reports that actual forest cover of India is 19.27% of the geographic area corresponding to 63.3 million hectare [1]. It is also estimated that the proportion of forest areas prone to forest fires annually ranges from 33% in some states to over 90% in others [1]. The forest fire season in India is usually from February to Mid June. An estimated annual economic loss of 400 Crore rupees is reported on account of forest fires over the country. Hence it is essential that research activities must be carried out for forest fire detection and extinguishing in order to save valuable resources, environment and economy of the country. Modeling fires is used to understand and predict fire behavior without getting burned [2]. In Antique days fire detection is carried out visibly by the forest guard who were walking along predefined paths also called ground patrolling and also by forest department personnel employed on the look out towers for keeping a vigil in the forest area through binoculars [3]. In both cases information is given to the control room and appropriate actions will be taken. Also both the methods are not reliable because the path may get damaged, bad weather conditions like mist will have poor visibility and life on the tower is miserable. In order to surmount the disadvantages faced by above methods some vision techniques such as Automatic Video Surveillance systems were proposed to monitor small forest. Some of the fire detection mechanisms such as AVHRR and MRIS which are based on satellite images are also proposed. But the long scan period and low resolution of images reduces the effectiveness of satellite based forest fire detection mechanisms because fire detection is very late and by that time fire is detected it may have grown large [4] and some of the of resources might have been destroyed. In order to overcome the difficulties associated with satellite image based mechanisms many works have been reported [5,6] using WSNs because sensory information can provide a more comprehensive forest fire monitoring with a finer grained spatial and temporal resolution. Also sensor nodes can be deployed in regions where there is no satellite signal. Most of the works reported in the literature are concerned with detection of fires based on simulation or using real test bed Environments [6]. Recent advancements in technology have lead to the emergence of distributed Wireless Sensor and Actor Networks [7]. In [8] forest fire detection and extinguishing using WSAN was proposed. In this paper we have presented a Grid based model based on matrix coordinates where each cell of a grid contains one sensor positioned at the center and a Path Planning algorithm for the Actor to extinguish fire. The remainder of the paper is organized as follows: section 2 describes Grid based model. In section 3 we present the methodology used for forest fire detection and extinguishing. Section 4 portrays the simulation results while section 5 concludes the paper.

————————————————

- *M.P.Sivaram kumar, Research Scholar, Department of Computer Science and Engineering, B S Abdur Rahman University, Vandalur, Chennai-600048.*
- *S.Rajasekaran, Professor, Department of Mathematics, B S Abdur Rahman University, Vandalur, Chennai-600048.*

## 2 GRID BASED MODEL

WSANs refer to a group of Sensors and Actors connected by wireless medium to perform distributed sensing and acting tasks. In WSANs the role of sensors is to gather information from the environment, while actors collect and process data and perform appropriate actions. The Notations and assumptions used in the grid based model is shown below

### 2.1 Notations and Assumptions

1. Define the forest area to be 'A' where the sensors are Initially deployed.
2. Consider a set of sensors S= {$s_1$, $s_2$, $s_3$…$s_n$} arranged in side a grid based on matrix coordinates in a two dimen -sional Euclidean region as shown in Fig 1.
3. Each Sensor is placed inside the forest area 'A' at coor -dinate ($x_i$,$y_i$) based on elements of a matrix and the sensor knows its own location.
4. The sensing region of a sensor is assumed to be circular with radius 'r', where 'r' is called as sensing range of sensor [9]
5. The length of a grid is equal to twice the amount of sensing range. i.e. 2r = L where 'L' is the length of the grid [9]
6. All the cells inside the grid are of equal size and have the same sensing range 'r'.
7. Each sensor has an Omni directional antenna. i.e. it can perform 360 degree observation.
8. Four sensor nodes are placed at the corners of the Do main.
9. Forest fire propagates in all directions at same speed.
10. Since propagation of fire is assumed to be uniform and we have considered surface fires only, fire origin -nated in a region i/cell i is detected by sensor i first.
11. Actor is available at the center of the Domain and re turns to the same place after it extinguishes fire.
12. Actor contains the processing unit to decide next se quence of moves to reach the target area and extin -guish fire.
13. Actors contain fire extinguishing powder to suppress fire.
14. Actor can move in any one of the eight directions [10] numbered 1, 2,3,4,5,6,7,8 from its current position as shown in Fig 2.
15. The obstacle positions are predefined.
16. Initial position of an Actor is represented using green oval and it is mobile.
17. Obstacles are represented using yellow color and the maximum length, width of the obstacle is equal to the side of square. i.e. obstacle is not greater than single square cell.
18. Obstacles are static.
19. A two dimensional array will hold a value of 1 for an obstacle in a cell whose coordinates are x, y. i.e. O[x, y] =1
20. The Path calculated by the Actor is shown using a Line in cyan color.

| (1,1) | (1,2) | (1,n-1) | (1,n) |
|---|---|---|---|
| (2,1) | (2,2) | (2,n-1) | (2,n) |
| . | . | . | . |
| (m-1,1) | (m-1,2) | (m-1,n-1) | (m-1,n) |
| (m,1) | (m,2) | (m,n-1) | (m,n) |

**Fig 1 Forest Area A divided into cells Based on Matrix Coordinates**

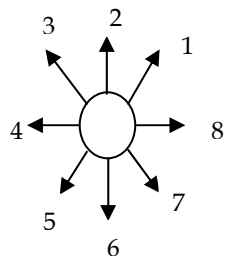

**Fig 2 Possible directions Actor can Move**

### 3. Forest Fire Extinguishing Model

In this work we have explored the use of WSANs in detecting and extinguishing forest fires. The automatic forest fire detection and extinguishing system consists of nodes deployed deterministically in a forest area and all the nodes know their location based on coordinate values of a matrix. Each node is equipped with a temperature sensor and an Omni directional antenna. Nodes continuously monitor the environment to check if there is fire or not in the particular cell. When change in temperature i.e. temperature raises above a certain threshold, is detected by a particular node they send message packets which contain location measurements. These packets are received by one of the corner node. The corner node then sends the packet to the Actor which in turn will process the packet which can be used in reaching the target area to extinguish fire.

Since the focus of this work is mainly on forest fire detection and extinguishing mechanism we assume that messages communicated has reached the corresponding destinations safely and if more than one node is sending the same message, the message sent by the first node is considered and rest of them are ignored. The Fundamental tasks to be considered in achieving this are classified into following

a. Sensor Deployment and Coverage
b. Communication between Nodes
c. Path planning for the Actor

### 3.1 Sensor Deployment and Coverage

According to [10] sensor's major function is to sense the environment for any incidence of event of interest. There-fore coverage is one of the major concerns in sensor networks. In [12], Coverage is classified into three classes: Area coverage, Point Coverage and Barrier coverage. Since our objective is to maximize the coverage percentage in this paper we focused on area coverage. The coverage problem is defined as how to position or deploy the sensors in a particular Region of Interest (ROI) so that coverage percentage is maximized and coverage holes are minimised. The deployment of nodes can be done either randomly or deterministically. In this work we have considered the deterministic deployment of nodes because sensor network coverage can be improved by carefully planning position of sensors in the ROI prior to their deployment [12]. Grid based sensor networks divide the ROI into square cells [8] and sensors can be placed at the center of the square cell in order to maximize the coverage and also the number of sensors required for placing inside the square cell is lesser than the number of sensors required for placing at the intersection of the grids [8]. Hence in this work we have placed the sensors at the center of the square cell. In case of Grid based deployment, problem of coverage of sensor field reduces to the problem of coverage of one cell and its neighbor because of symmetry of cells. Therefore if we take one cell and calculate the uncovered area it is approximately equal to $0.86 \, r^2$ [13] where 'r' is the sensing range. In this work we have assumed that fire spreads at a point and if it occurs inside the uncovered area of sensor i, then it will be detected by sensor i first rather than any other sensor because surface fire spreads at a constant rate in all directions [8].

### 3.2 Communication between Nodes

In this work we have used three nodes (i) sensor nodes capable of detecting a change in temperature which indicates the fire occurrence and it broadcasts the message to the neighboring cells (ii) Corner nodes receives the message sent by the sensor nodes and it has got high capacity for transmission of messages (iii) Actor node receives the information from corner node the coordinates of the cell where the fire has occurred. Then with its own coordinate as start position and received coordinates as goal position it plans a path along which it will travel and reach the goal position and start extinguishing fire. When more than one corner node is sending the same coordinate information regarding fire it will be discarded. In these works we have assumed single point of occurrence of fire in the entire Domain. Therefore there is no possibility of different corner nodes sending different coordinates for fire occurrence. Also we assume that the messages sent from one node to another node will reach without loss.

### 3.3 Path Planning for the Actor

The path panning for mobile robots is defined as the search for a path which a robot has to follow in a predefined environment in order to reach a particular position [14]. Path planning is considered as an optimization problem of finding a path between start and end points that should be free of collision and the path obtained is the shortest one [15]. Mobile robot path planning can be classified as (i) static or dynamic (ii) local or global and (iii) complete or heuristic based on environment, algorithm and completeness. Static path planning refers to the environment which does not contain moving objects and obstacles. Dynamic path planning refers to the environment which contains moving objects and obstacles. Both static and dynamic path planning contains mobile robot. When the robot knows the information about the environment the path planning is said to be global. On the contrary if the robot does not know the information about the environment then the path planning is local. In this work the path planning of actor to reach the target area and extinguish fire is simulated using the actions performed by a robot. In our work we have considered path planning for two types of environment (i) Environment without obstacles. This type of environment is similar to savannah type forest where there are grasslands only available and no obstacles. (ii) Environment with obstacles. This is similar to ordinary forest where there are trees, stones and some other static obstacles. The path planning of the Actor is done in incremental steps by taking into consideration the start position coordinates and goal position coordinates. Since in this work we are using matrix based coordinates for the grid, the determination of direction of in which the actor has to move presented in [16] is modified and it is shown below

| S.No | Condition of the coordinates | Direction it has to move |
|---|---|---|
| 1. | $(x_s-x_g=0)$ and $(y_s-y_g<0)$ | Right |
| 2. | $(x_s-x_g>0)$ and $(y_s-y_g)<0)$ | Diagonal Up Right |
| 3. | $(x_s-x_g>0)$ and $(y_s-y_g)=0)$ | Up |
| 4. | $(x_s-x_g>0)$ and $(y_s-y_g)>0)$ | Diagonal Up Left |
| 5. | $(x_s-x_g=0)$ and $(y_s-y_g)>0)$ | Left |
| 6. | $(x_s-x_g<0)$ and $(y_s-y_g)>0)$ | Diagonal Down Right |
| 7. | $(x_s-x_g<0)$ and $(y_s-y_g)=0)$ | Down |
| 8. | $(x_s-x_g<0)$ and $(y_s-y_g)<0)$ | Diagonal Down Left |

### 3.3.1 Path planning Algorithm for Environment without obstacles

Step1: Read the $x_g$ coordinate and $y_g$ coordinate from the message received from corner node.

Step2: Since we know that Actor will always be available at the center assign the coordinate values of actor to $x_s$ and $y_s$.

Step3: Repeat until ($x_s=x_g$ and $y_s=y_g$)
    if (($x_s-x_g=0$) and ($y_s-y_g<0$)) then
      Move Right
    elseif (($x_s-x_g>0$) and ($y_s-y_g)<0$)) then
      Move Diagonal Up Right

```
            elseif ((xs-xg>0) and (ys-yg)=0)) then
                Move Up
            elseif ((xs-xg>0) and (ys-yg)>0)) then
                Move Diagonal Up Left
            elseif ((xs-xg=0) and (ys-yg)<0)) then
                Move Left
            elseif ((xs-xg<0) and (ys-yg)>0)) then
                Move Diagonal Down Right
            elseif ((xs-xg<0) and (ys-yg)=0)) then
                Move Down
            elseif ((xs-xg<0) and (ys-yg)<0)) then
                Move Diagonal Down Left
            end if
        end if
      end if
     end if
    end if
   end
Step4: Stop the Actor and start extinguish fire.
```

### 3.3.1 Path planning Algorithm for Environment with obstacles

Step1: Read the xg coordinate and yg coordinate from the message received from corner node.

Step2: Since we know that Actor will always be available at the center assign the coordinate values of actor to xs and ys.

Step3: do
```
      {
         d= direction (xs,xg,ys,yg)
         switch (d)
         {
           Case 1: if (O[xs-1,ys+1] !=1 ) then
                    Move Diagonal Up Right
                   else
                    Call freespacesearch(xs,ys,1)
                   end if
                   break
           Case 2: if (O[xs-1,ys] !=1 ) then
                    Move Up
                   else
                    Call freespacesearch(xs,ys,2)
                   end if
                   break
           Case 3: if (O[xs-1,ys-1] !=1 ) then
                    Move Diagonal Up Left
                   else
                    Call freespacesearch(xs,ys,3)
                   end if
                   break
           Case 4: if (O[xs,ys-1] !=1 ) then
                    Move Left
                   else
                    Call freespacesearch(xs,ys,4)
                   end if
                   break
           Case 5: if (O[xs+1,ys-1] !=1 ) then
                    Move Diagonal Down Right
                   else
                    Call freespacesearch(xs,ys,5)
                   end if
                   break
           Case 6: if (O[xs+1,ys] !=1 ) then
                    Move Down
                   else
                    Call freespacesearch(xs,ys,6)
                   end if
                   break
           Case 7: if (O[xs+1 ,ys+1] !=1 ) then
                    Move Diagonal Down Left
                   else
                    Call freespacesearch(xs,ys,7)
                   end if
                   break
           Case 8: if (O[xs,ys+1]!=1 ) then
                    Move Right
                   else
                    Call freespacesearch(xs,ys,8)
                   end if
                   break
         }
      }while((xs – xg != 0) and (ys – yg !=0))
Step4: Stop the Actor and start extinguish fire.
```

### 3.3.2.1 Procedure direction

This procedure is used to determine the direction the actor has to move based on start and goal coordinates. It will return an integer value based on the direction it has to move.

int direction($x_s, x_g, y_s, y_g$)
```
{
       if ((xs-xg>0) and (ys-yg)<0)) then
            return 1
       elseif ((xs-xg>0) and (ys-yg)=0)) then
            return 2
       elseif ((xs-xg>0) and (ys-yg)>0)) then
            return 3
       elseif ((xs-xg=0) and (ys-yg)>0)) then
            return 4
       elseif ((xs-xg<0) and (ys-yg)>0)) then
            return 5
       elseif ((xs-xg<0) and (ys-yg)=0)) then
            return 6
       elseif ((xs-xg<0) and (ys-yg)<0)) then
            return 7
       elseif ((xs-xg =0) and (ys-yg < 0 )) then
            return 8
       end if
    end if
```

```
        end if
        end if
        end if
        end if
        end
}
```

### 3.3.2.1 Procedure freespacesearch

This procedure is used to determine the cell which does not contain an obstacle if the next move of the actor to the cell contains an obstacle. It takes three arguments the s which refers to x coordinate, f which refers to y coordinate and e is the direction from the current cell in which check for obstacle is made and cell contained an obstacle. Therefore we perform free space search i.e. finding the next cell which is free without any obstacle in the anti clock wise direction using the procedure shown below.

```
freespacesearch(s,f,e)
{
   switch (t)
   {
        Case 1: if (O[x_s-1,y_s+1] !=1 ) then
                x_s=x_s-1
                y_s=y_s+1
                Move to x_s,y_s
                return
                break
        Case 2: if (O[x_s-1,y_s] !=1 ) then
                x_s=x_s-1
                Move to x_s,y_s
                return
                break
        Case 3: if (O[x_s-1,y_s-1] !=1 ) then
                x_s=x_s-1
                y_s=y_s-1
                return
                break
        Case 4: if (O[x_s,y_s-1] !=1 ) then
                ys=ys-1
                Move to x_s,y_s
                return
                break
        Case 5: if (O[x_s+1,y_s-1] !=1 ) then
                x_s=x_s+1
                y_s=y_s-1
                Move to x_s,y_s
                return
                break
        Case 6: if (O[x_s+1,y_s] !=1 ) then
                x_s=x_s+1
                Move to x_s,y_s
                return
                break
        Case 7: if (O[x_s+1 ,y_s+1] !=1 ) then
                x_s=x_s+1
                y_s=y_s+1
                Move to x_s,y_s
                return
                break
        Case 8: if (O[x_s,y_s+1] !=1 ) then
                y_s=y_s+1
                Move to x_s,y_s
                return
                break

        Default : No path for the robot to move
                exit
   }
}
```

## 4. Simulation results

### 4.1 Simulation in an Environment without obstacle

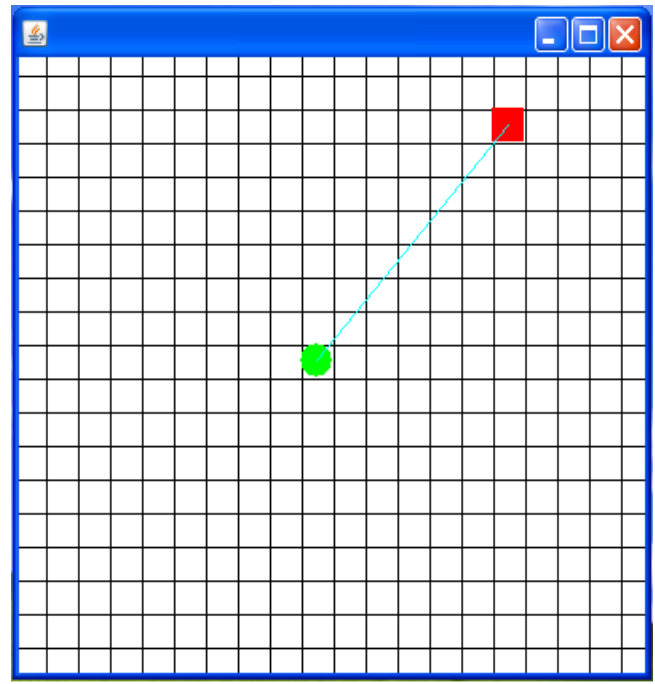

In this work we have proposed a grid containing m x n cells based on matrix coordinates as a forest domain in which fire occurrence is automatically detected by sensors and extinguished by actors. In the simulation part we have not shown the corner nodes because we have assumed that message sent by sensors to corner node and from corner node to actor is reached without any loss since the prime consideration of this work is extinguishing of forest fires. Also we have considered a single point of occurrence of fire since we have only one actor for the entire forest domain. We have devised a path planning algorithm for the actor in order to extinguish fire in both types of environment.i.e. Environment with and without obstacles. We have used java for simulation purposes. We created a 20 x 20 grid. The actor is represented by a green oval located at the center, the cells containing obstacles are represented using yellow and the cell where fire occurs is shown in red and a line in cyan color shows the path planning of the Actor to travel and reach the target area to extinguish fire. To test the effectiveness of the algorithm we have proposed, we created fire in various cells including all the quadrant regions and horizontal and vertical lines by varying the number of obstacles. In all the cases the path is created using the algorithm without any collision with the obstacle and the actor travels through the path to extinguish fire. The simulation results are shown below.

### 4.2 Simulation in an Environment with obstacle

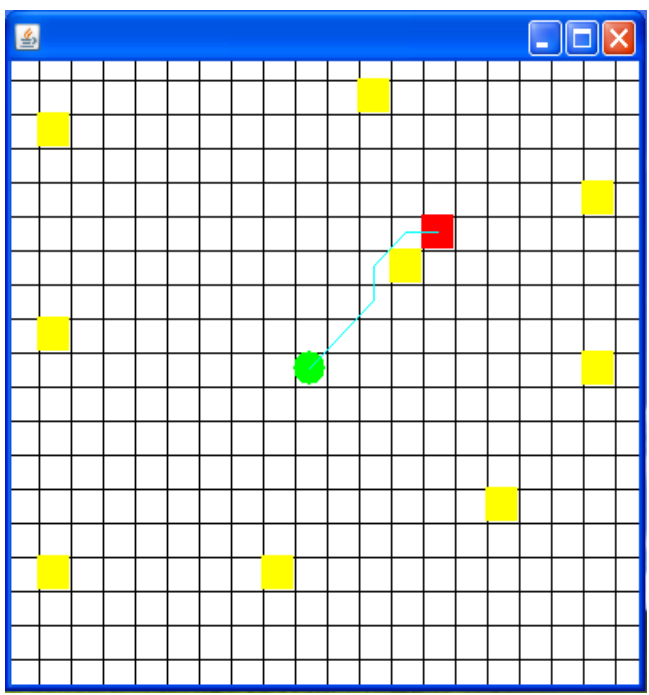

## 5. Conclusion

In this paper we have presented a path planning algorithm for the actor to move to the target area to extinguish fire. In this work we have assumed that robot can move in all directions without any rotation of robot. i.e. if the robot is facing towards east and the next movement it has to make is towards north then robot has to do some adjustment so that it can turn towards north first and then move towards the cell. We have considered only 9 obstacles for 20 x 20 environments. This simulation will work for any number of obstacles placed in any position. In this work we have considered only static obstacles but in future we have plans to incorporate as many dynamic obstacles as possible and also consider time taken for the robot to move towards the target area and also the time taken for the rotation of robot in order to perform next move.

## 6. References

[1] Http: //www.fire.uni-freiburg.de
[2] Qasim Siddque, "Survey of Forest Fire Simulation", Global journal of computer science and technology, pp.137-148
[3] Pablo.I.Fierens, "Number of Wireless sensors needed to detect a wildfire", Informal Publication
[4] M.Hefeeda, M.Bagheri, "Forest Fire Modeling and Early Detection using Wireless Sensor Networks" Ad Hoc & Sensor Wireless Networks Vol.7, pp.169-224
[5] Zhang J G, Li W B, K J M, "Forest fire detection system based on ZigBee wireless sensor network [J] ,Journal of Beijing Forestry University ,v 29,n 4,July,2007,pp.41-45
[6] B.Kosucu, K.Irgan, G.Kucuk, S.Bay dere, "Fire Sense TB: A wireless sensor networks Test Bed for Forest Fire Detection", pp.1173-1177
[7] Ian F.Akyildiz, Ismail H.Kasimoglu, "Wireless Sensor and Actor Networks: research Challenges", Mobile Adhoc Networks, Vol.no 2, pp.351-367
[8] M.P.Sivaram Kumar, S.Rajasekaran, "Detection and Extinguish-ing Forest Fires using Wireless Sensor and Actor Networks", International Journal of Computer Application,Vol.24 (1), June 2011, pp.31-35.
[9] B Liu, D Towsley, "On the coverage and detectability of large scale wireless sensor networks" proceedings of the workshop on Modeling and optimization in Mobile Adhoc and wireless networks (Wiopr'03), Mar 2003
[10] M Seda and T Brezina, "Robot Motion Planning in Eight Directions", Proceedings of the World Congress on Engineering and Computer Science 2009 Vol II WCECS 2009, October 2009, San Francisco, USA
[11]N.A. Ab Aziz, K.Ab Aziz and W.Z Wan Ismail, "Coverage Strategies for Wireless Sensor Networks", World Academy of Science, Engineering and Technology 50 2009, pp.145-150
[12] Cardei M and Wu J, "Coverage in Wireless Sensor Networks" Informal Publication
[13] Jiehuichen, M B Salim and M Matsumoto, "A single mobile target tracking in voronoi based clustered wireless sensor network", Journal of Information processing systems Vol.7, March 2011, pp.17-28
[14] Buniyamin N, Wan Ngah W A J, Sariff N, Mohammad Z, "A Simple Local Path Planning Algorithm for Autonomous Mobile robots", International Journal of systems applications, Engineering & development issue 2, volume 6, pp 151-159, 2011
[15] A.howard, M.J.Matari and G.S.Sukhatme, "An Incremental self deployment algorithm for Mobile sensor Networks, autonomous robots", special issue on Intelligent Embedded systems, 13(2), September 2002, pp.113-126.
[16] Ting-Kai Wang, Quan Dang, Pei-Yuan Pan, "Path Planning Approach in Unknown Environment", International Journal of Automation and Computing", 7(3), August 2010, pp.310-316.